\documentclass[letterpaper, 10 pt, conference]{ieeeconf}  

\IEEEoverridecommandlockouts                              

\usepackage{tikz}

\def\*#1{\mathbf{#1}}

\usepackage{amsmath}
\usepackage{amsfonts}
\usepackage{dsfont}

\usepackage{booktabs} 
\usepackage{multirow}
\usepackage{graphicx}
\usepackage{cleveref}

\usepackage{amssymb}
\usepackage{pifont}

\usepackage{siunitx}

\title{\LARGE \bf
Towards Robust 3D Object Detection In Rainy Conditions
}

\author{Aldi Piroli$^{1}$, Vinzenz Dallabetta$^{2}$, Johannes Kopp$^{1}$, Marc Walessa$^{2}$, \\ Daniel Meissner$^{2}$, and Klaus Dietmayer$^{1}$
\thanks{$^{1}$ Institute of Measurement, Control, and Microtechnology, Ulm University, Germany {\tt\small \{firstname.lastname\}@uni-ulm.de}}
\thanks{$^{2}$ BMW~AG, Petuelring 130, 80809~Munich,~Germany {\tt\small \{vinzenz.dallabetta, marc.walessa\}@bmw.de} and {\tt\small daniel.da.meissner@bmwgroup.com}}%
}

\newcommand\copyrighttext{%
	\footnotesize \copyright\,2023 IEEE. Personal use of this material is permitted. Permission from IEEE must be obtained for all other uses, in any current or future media, including reprinting/republishing this material for advertising or promotional purposes, creating new collective works, for resale or redistribution to servers or lists, or reuse of any copyrighted component of this work in other works.}%
\newcommand\copyrightnotice{%
	\begin{tikzpicture}[remember picture,overlay]%
	\node[anchor=south,yshift=10pt] at (current page.south) {\fbox{\parbox{\dimexpr\textwidth-2cm}{\copyrighttext}}};%
	\end{tikzpicture}%
	\vspace{-10pt}%
}

\begin{document}

\maketitle
\copyrightnotice
\thispagestyle{empty}
\pagestyle{empty}

\begin{abstract}
LiDAR sensors are used in autonomous driving applications to accurately perceive the environment.
However, they are affected by adverse weather conditions such as snow, fog, and rain.
These everyday phenomena introduce unwanted noise into the measurements, severely degrading the performance of LiDAR-based perception systems.
In this work, we propose a framework for improving the robustness of LiDAR-based 3D object detectors against road spray.
Our approach uses a state-of-the-art adverse weather detection network to filter out spray from the LiDAR point cloud, which is then used as input for the object detector.
In this way, the detected objects are less affected by the adverse weather in the scene, resulting in a more accurate perception of the environment. 
In addition to adverse weather filtering, we explore the use of radar targets to further filter false positive detections.
Tests on real-world data show that our approach improves the robustness to road spray of several popular 3D object detectors.
\end{abstract}

\section{Introduction}
\begin{figure}[t!]
    \centering
    \includegraphics[width=\columnwidth]{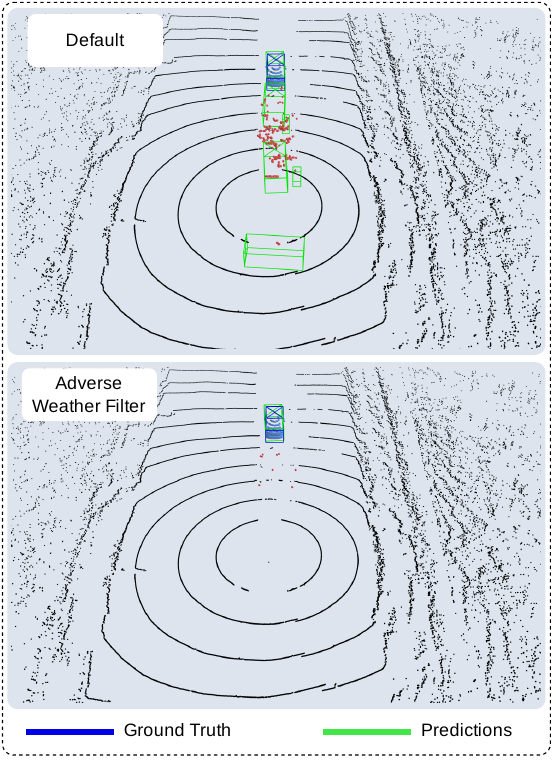}
    \caption{
The performance of LiDAR-based perception systems is seen to degrade in adverse weather conditions. 
In the top figure, we show the effect of road spray on the object detector SECOND~\cite{yan2018second},  trained on good weather conditions.
The presence of road spray causes the detector to predict ghost objects.
In the bottom figure, we see that filtering out adverse weather points from the input point cloud improves the overall quality of the detections.
In the figures, we show spray points in red and vehicle points in light blue.
}     \label{Fig:teaser}
\end{figure}

Autonomous vehicles rely on camera, LiDAR, and radar sensors to perceive the environment.
Compared to the other sensors, LiDARs offer rich depth information regardless of the lighting conditions.
LiDAR-based perception systems like semantic segmentation and 3D object detection perform extremely well in good weather conditions.
However, their performance is seen to degrade when testing these models in adverse weather like snow, fog, and rain~\cite{Piroli2023EnergybasedDO, mirza2021robustness,Piroli2022DetectionOC, Dong_2023_CVPR, Piroli2022Robust3O, HahnerICCV21, hahner2022lidar}.
In this paper, we will focus on the effect of vehicle road spray on 3D object detection.
The spray effect is commonly observed when a vehicle is traveling at high speeds on a wet road surface.
The forward movement of the vehicle's tires causes the water particles on the ground to be propelled behind the vehicle, creating a so-called spray corridor.
These particles are detected by LiDAR sensors and can be seen as a form of unwanted noise in the measurement.
Object detectors that are trained on good weather conditions are negatively impacted by this phenomenon.
The resulting effects are the missed detection of objects due to the blocked field of view and the introduction of ghost objects (false positive detections) into the perception system~\cite{walz2021benchmark}.
This last effect can be extremely problematic in everyday scenarios.
For example, while traveling on a highway, spray could be detected as an oncoming vehicle and, in extreme cases, cause the autonomous vehicle to perform an evasive maneuver, endangering its passengers and other road users. 
An example of ghost object detections is shown in Fig.~\ref{Fig:teaser}.
In the literature, few works have been proposed to improve the robustness of object detectors in adverse weather.
Hahner et al.~\cite{hahner2022lidar} propose simulating scenes in snowy conditions to train object detectors.
Xu et al.~\cite{xu2021spg} upsample objects in a point cloud in rainy conditions to account for the lower number of returned points.
However, no explicit solution is proposed to deal with the unwanted measurement noise caused by spray.
Linnhoff et al.~\cite{linnhoff2022simulating} present a method to simulate road spray and use this data to improve the robustness of object detection.
However, this requires retraining the object detector, which can be expensive or even infeasible in the case of commercially available detectors.
In addition, introducing noise into the training data could lead to lower performance when tested on good weather data~\cite{Piroli2022Robust3O} and the simulation models may not fully reflect the spray generated in real and complex scenarios.

In this paper, we present a simple yet effective framework for improving the robustness of object detectors in rainy conditions.
Our framework uses a state-of-the-art adverse weather detection network~\cite{Piroli2023EnergybasedDO} to identify and remove spray points in a LiDAR point cloud.
The filtered data is then used as input to an object detector trained only on good weather data.
As many autonomous driving application systems include both LiDAR and radar sensors, we explore the use of radar targets as an additional post-processing step to filter out ghost detections. 
We test our framework on the SemanticSpray dataset~\cite{Piroli2023EnergybasedDO}, which contains highway-like scenarios in rainy conditions, and show that it improves the robustness of the evaluated object detectors to spray.
Furthermore, since we do not require the re-training of the object detectors, performance in good weather conditions remains unchanged. 

In summary, our main contributions are:
\begin{itemize}
  \item We propose a framework for improving the robustness of 3D LiDAR object detectors in rainy conditions, based on point-wise adverse weather filtering.
  \item For multimodal setups that include both LiDAR and radar sensors, we employ radar targets to further filter out ghost objects caused by spray.   
  \item Tests on real-world scenarios show that our method improves the robustness of multiple object detectors to road spray, while still allowing for real-time performance. 
\end{itemize}

\section{Related Work}
\subsection{3D Object Detection on Point Clouds}
The goal of 3D object detectors is to return a set of 3D bounding boxes containing the relevant objects in the scene.
A variety of approaches have been proposed to solve this problem.
Early methods like VoxelNet~\cite{zhou2018voxelnet} use an intermediate voxel representation for the input point cloud, and then extract features using full 3D convolutions.
SECOND~\cite{yan2018second} improves the voxel-based architecture by using sparse convolutions, greatly reducing computation times.
PointPillars~\cite{lang2019pointpillars} further reduces computation by projecting the point cloud in a 2D pillar-based structure and uses 2D convolutions to extract features.
Recently, center-based approaches like CenterPoint~\cite{yin2021center} have been proposed, which allow for the detection of objects without the use of predefined anchors, making detections more robust to orientation changes. 
State-of-the-art networks like VoxelNeXt~\cite{chen2023voxenext} use instead only 3D sparse voxel features to predict objects without the need for anchor or center proxies.

\subsection{LiDAR Perception in Adverse Weather}
LiDAR sensors are negatively affected by adverse weather effects like snow, fog, and rain~\cite{walz2021benchmark, mirza2021robustness, Dong_2023_CVPR}. 
The effect of spray on autonomous driving perception systems is examined by Walz et al.~\cite{walz2021benchmark}.
Using a vehicle-mounted device to simulate trailing spray, they find that both LiDAR and camera-based detectors are impacted by this effect, which significantly reduces their reliability.
For LiDAR-based detectors, they observe both the introduction of ghost objects caused by spray points and the missed detection of the leading vehicle due to the obstructed field of view. 
Different approaches exist for detecting adverse weather effects in LiDAR point clouds.
Charron et al.~\cite{charron2018noising} propose the DROR filter, which aims to remove snow points in a point cloud by using local neighbor information to determine whether a point is an outlier or not.
Kurup et al.~\cite{kurup2021dsor} improve on this concept by proposing the DSOR filter, which incorporates the mean distance between neighbors during filtering.
Piroli et al.~\cite{Piroli2022DetectionOC}  detected vehicle exhaust  by identifying the possible emission area for each vehicle in the scene and then finding regions in the point cloud where exhaust clouds are likely to be present. 
Heinzler et al.~\cite{heinzler2020cnn} use a weather chamber to generate artificial fog and rain, and then use a lightweight CNN network to classify the LiDAR points associated with adverse weather.
Stanislas et al.~\cite{stanislas2021airborne} propose both a voxel-based and a CNN-based architecture to identify airborne particles such as dust and smoke in LiDAR point clouds.   
In a recent work of ours, we propose AWNet~\cite{Piroli2023EnergybasedDO}, which uses an energy-based outlier detection framework to distinguish between inliers (non-adverse weather) and outliers (adverse weather).
AWNet achieves state-of-the-art performance in snow, rain, fog, and spray detection.
It also shows promising results in generalizing to unseen weather effects.

Few works have been proposed to improve the robustness of object detection in adverse weather conditions.
Bijelic et al.~\cite{bijelic2020seeing} propose a single-shot model that fuses camera, LiDAR, and radar information to robustly detect objects in foggy scenes.
Hahner et al.~\cite{HahnerICCV21,hahner2022lidar}  use  simulated fog and snowfall to improve the robustness of object detectors. 
Recently, Linnhoff et al.~\cite{linnhoff2022simulating} propose a simulation model for road spray effects and then use the data to improve object detection.
In contrast, our framework first removes the spray points and then uses the filtered point cloud as input to the detector.
This allows detectors that have only been trained on good weather data to be used in rainy conditions without the need for retraining.
Furthermore, our proposed filtering approach can be activated or deactivated based on the contextual perception of the environment (e.g., activated only when it is raining)~\cite{9748024}.
\begin{figure*}[t!]
    \centering
        \includegraphics[width=0.95\textwidth]{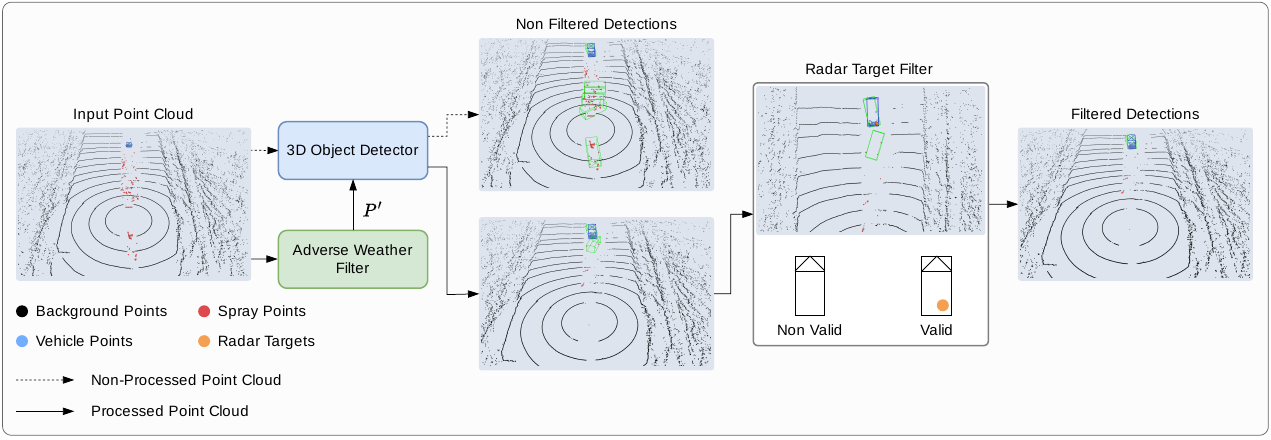}
    \caption{
Overview of our proposed method. 
Given a LiDAR point cloud $\*P$, we first remove all adverse weather points and obtain a filtered point cloud $\*P'$ which is used as input for the object detector. 
Compared to directly using the unprocessed point cloud as input, the resulting detections are less affected by road spray. 
In a multi-sensor setup that includes radar, we use its inherent robustness to adverse weather to further filter out false positive detections by checking if a detected object has an associated radar target.
The point colors are used for visualization purposes only.
We use blue boxes to represent the ground truth and green for object detection.}    \label{Fig:method}
\end{figure*}
\section{Method}
In the following, we describe our framework for improving the robustness of an object detector against road spray.
Our method can be used in a single-modality application (LiDAR sensor only) or in a multimodal approach (LiDAR and radar sensors).
An overview of the method is given in Fig.~\ref{Fig:method}.

\subsection{Sensor Input Data}
As primary sensor data, we use a LiDAR point cloud $\*P\in\mathbb{R}^{N \times C}$ composed of $N$ points each with $C$ features (i.e., $x,y,z,\text{\textit{intensity}}$).
For the multimodal approach, we  assume to have a radar target list $\*R\in\mathbb{R}^{M \times 4}$, with each target defined by its position and velocity (i.e., $x,y,z,v$).
Both LiDAR and radar sensors are assumed to be time synchronized and calibrated to the same reference frame.

\subsection{Adverse Weather Filtering }
Given a LiDAR point cloud $\*P$, we want to detect and remove all points caused by road spray and obtain a filtered point cloud $\*P'\in \mathbb{R}^{N' \times C}$, with $N' \leq N$.
To this end, we employ a network $\phi :\mathbb{R}^{N \times C} \rightarrow \mathbb{R}^{N}$ which takes $\*P$ as input and classifies all of the $N$ points as \textit{valid} or \textit{noise}.
As the main method for $\phi$, we use AWNet~\cite{Piroli2023EnergybasedDO}, a state-of-the-art network for detecting adverse weather points in LiDAR point clouds.
AWNet detects adverse weather by returning an anomaly score for each point in $\*P$. 
More specifically, AWNet learns to detect adverse weather points as outliers and uses the point energy score~\cite{Piroli2023EnergybasedDO,liu2020energy} to determine if a point is an inlier or not. 
A binary classification can be performed by selecting an appropriate  threshold $\tau$ common for each point, so that 
\begin{align}
\label{eq:decision_rule}
\begin{cases}
\text{\textit{valid}} & \quad \text{if } \phi(\*P)_i \leq \tau, \\
\text{\textit{noise}}  & \quad \text{else},
\end{cases}
\end{align}
where $\phi(\*P)_i$ is the output corresponding to the $i$-th point of $\*P$.
Using this decision rule, the filtered point cloud $\*P'$ is obtained by selecting all points of $\*P$ classified as \textit{valid}.

\begin{table*}[t!]
    \centering
    \caption{Evaluation results of our framework with different modalities activated.
    The results refer to the 3D AP metric measured at different distances and on the overall detection range. 
    The symbol \dag~refers to the single modality method that only use the LiDAR sensor.
    Instead, the symbol \ddag~refers to multimodal methods that also use radar data.
        Values are in percentage. Bold numbers represent the best results. }
    \resizebox{0.8\textwidth}{!}{%
        \begin{tabular}{@{}llcccccc@{}}
            \toprule
            \multirow{2}{*}{\textbf{Processing}}                                                                     & \multirow{2}{*}{\textbf{Detector}} & \multicolumn{3}{c}{\textbf{3D AP w/o fine-tuning}} & \multicolumn{3}{c}{\textbf{3D AP w/ fine-tuning}}                                                                                       \\ \cmidrule(l){3-8}
                                                                                                                     &                                    & \textbf{$0$-\SI{25}{\meter}}                       & $>\SI{25}{\meter}$                                & overall        & \textbf{$0$-\SI{25}{\meter}} & $>\SI{25}{\meter}$ & overall        \\ \midrule
            \multirow{4}{*}{\dag~No Preprocessing}                                                                        & PointPillars                        & 69.42                                              & 40.06                                             & 57.20          & 92.88                        & 76.85              & 86.00          \\
                                                                                                                     & SECOND                             & 98.22                                              & 84.23                                             & 92.39          & 97.70                        & 87.50              & 93.58          \\
                                                                                                                     & CenterPoint                        & 73.79                                              & 11.95                                             & 41.74          & 93.88                        & 77.21              & 86.80          \\ \cmidrule(l){2-8}
                                                                                                                     & \textit{average}                   & 80.48                                              & 45.41                                             & 63.78          & 94.82                        & 80.52              & 88.79          \\ \midrule
            \multirow{4}{*}{\begin{tabular}[l]{@{}l@{}}\dag~Adverse Weather Filter \\ using AWNet~\cite{Piroli2023EnergybasedDO}\end{tabular}}           & PointPillars                        & 83.23                                              & 43.04                                             & 68.53          & 95.02                        & 77.20              & 88.64          \\
                                                                                                                     & SECOND                             & 99.37                                              & 84.86                                             & 93.60          & 99.43                        & 85.64              & 94.83          \\
                                                                                                                     & CenterPoint                        & 81.37                                              & 12.95                                             & 46.23          & 96.15                        & 79.38              & 90.11          \\ \cmidrule(l){2-8}
                                                                                                                     & \textit{average}                   & \textbf{87.99}                                     & \textbf{46.95}                                    & \textbf{69.45} & \textbf{96.87}               & \textbf{80.74}     & \textbf{91.19} \\ \midrule \midrule
            \multirow{4}{*}{\ddag~Radar Target Filter}                                                                     & PointPillars                        & 74.50                                              & 41.38                                             & 60.46          & 95.54                        & 78.62              & 88.81          \\
                                                                                                                     & SECOND                             & 99.55                                              & 85.97                                             & 94.62          & 99.66                        & 86.29              & 95.19          \\
                                                                                                                     & CenterPoint                        & 97.50                                              & 73.20                                             & 89.78          & 97.61                        & 82.62              & 92.84          \\ \cmidrule(l){2-8}
                                                                                                                     & \textit{average}                   & 90.52                                              & 66.85                                             & 81.62          & 97.60                        & 82.51              & 92.28          \\ \midrule
            \multirow{4}{*}{\begin{tabular}[l]{@{}l@{}}\ddag~Adverse Weather Filter + \\ Radar Target Filter\end{tabular}} & PointPillars                        & 83.72                                              & 43.76                                             & 67.53          & 95.63                        & 78.79              & 88.93          \\
                                                                                                                     & SECOND                             & 99.54                                              & 86.20                                             & 94.73          & 99.64                        & 86.02              & 95.11          \\
                                                                                                                     & CenterPoint                        & 99.51                                              & 73.29                                             & 90.01          & 97.60                        & 82.88              & 92.93          \\ \cmidrule(l){2-8}
                                                                                                                     & \textit{average}                   & \textbf{94.26}                                     & \textbf{67.75}                                    & \textbf{84.09} & \textbf{97.62}               & \textbf{82.56}     & \textbf{92.32} \\ \bottomrule
        \end{tabular}
    }
    \label{tab:weather_filter_and_radar}
\end{table*}

\subsection{Object Detection}
A 3D object detector $\psi:\in\mathbb{R}^{N' \times C} \rightarrow \mathbb{R}^{D \times 7}$ is used to detect the relevant objects in the scene.
The detector takes as input the filtered point cloud $\*P'$ and returns a set of object detections $\mathcal{D}$.
Each detection  $d \in \mathcal{D}$ is characterized by a bounding box $\left[x,y,z,w,l,h,\theta \right]$, which describes the object's position, size, and orientation.
By using $\phi$ to filter out noise in the point cloud, the object detector $\psi$ is less likely to return unwanted ghost object detections caused by the vehicle's trailing spray, resulting in a more robust perception of the environment. 

\subsection{Radar-based Detection Filtering}
In addition to the LiDAR-only filtering approach described in the previous section, we also show how radar targets can be used to improve LiDAR-based 3D object detections. 
For each detection $d \in \mathcal{D}$ we check if one or more radar targets are contained in the object bounding box.
To account for errors (e.g., size and orientation) in the estimated bounding boxes, we increase the dimension of each bounding box by adding a fixed padding $\gamma$, so that the bounding box associated with an object is equal to $\left[x,y,z,w+\gamma,l+\gamma,h+\gamma,\theta \right]$.
As radar sensors are inherently robust to adverse weather conditions, detections caused by spray will have no associated radar targets and can therefore be filtered out using this simple rule.

\section{Experiments}
\subsection{Experiment Setup}
\textbf{Datasets.}
For the training of the object detectors, we use the NuScenes dataset~\cite{caesar2020nuscenes}, which contains approximately $\SI{40}{\kilo{}}$ scans recorded a mix of urban and rural scenarios.
Although some of the scenes are recorded in adverse weather (e.g., snow, rain) no large trailing spray effect is present.
To evaluate the effectiveness of our framework in improving object detection in rainy conditions, we use the recently released SemanticSpray dataset~\cite{Piroli2023EnergybasedDO}, which provides semantic labels for a subset of scenes of the Road Spray dataset~\cite{road_spray_dataset}.
The dataset provides a large-scale study of motorway-like scenarios where both the ego and a leading vehicle travel through a wet surface, creating a trailing spray effect.
Multiple scenarios are present with vehicles traveling at different speeds (from \SI{50}{\kilo\meter\per\hour} to \SI{130}{\km\per\hour}) and various levels of water on the road surface.
Since no ground truth 3D box annotations are available, we use an automated pipeline to generate ground truth data.
Similar to~\cite{linnhoff2022simulating}, we generate object detections using a high-performing 3D object detector (in our case  VoxelNeXt~\cite{chen2023voxenext}).
Afterward, we use an object-tracking framework~\cite{3Dtracker} to improve the initial predictions.
Finally, we use the semantic labels to filter out the predicted bounding boxes that are not generated by vehicles, labeling a total of $4665$ scenes.

\textbf{Adverse Weather Effect Detection Methods.}
We follow our previous work~\cite{Piroli2023EnergybasedDO} for all of the adverse weather detection methods.
We use AWNet~\cite{Piroli2023EnergybasedDO} as the primary method for spray detection and additionally provide the results when using other adverse weather detection methods, namely Particle-VoxelNet~\cite{stanislas2021airborne}, Particle-UNet~\cite{stanislas2021airborne}, WeatherNet~\cite{heinzler2020cnn}, and DSOR~\cite{kurup2021dsor}.
All learning-based methods are trained on the SemanticSpray dataset.
For details on the implementation and training of each method, we refer the reader to~\cite{Piroli2023EnergybasedDO}.
In all experiments, unless otherwise stated, we use for AWNet a threshold $\tau$ at $99\%$ TPR (true positive rate), and additional bounding box padding $\gamma = \SI{1.0}{\meter}$.

\begin{table*}[t!]
    \centering
    \caption{Evaluation results of different adverse weather detection methods.
        The results refer to the 3D AP metric measured at different distances and on the overall detection range. 
       Values are in percentage. Bold numbers represent the best results.}
    \resizebox{0.8\textwidth}{!}{%
        \begin{tabular}{@{}llcccccc@{}}
            \toprule
            \multirow{2}{*}{\textbf{Filter}}   & \multirow{2}{*}{\textbf{Detector}} & \multicolumn{3}{c}{\textbf{3D AP w/o fine-tuning}} & \multicolumn{3}{c}{\textbf{3D AP w/ fine-tuning}}                                                                              \\ \cmidrule(l){3-8}
                                               &                                    & $0$-\SI{25}{\meter}                                & $>\SI{25}{\meter}$                                & overall        & $0$-\SI{25}{\meter} & $>\SI{25}{\meter}$ & overall        \\ \midrule
            \multirow{4}{*}{No Preprocessing}  & PointPillars                        & 69.42                                              & 40.06                                             & 57.20          & 92.88               & 76.85              & 86.00          \\
                                               & SECONDNet                          & 98.22                                              & 84.23                                             & 92.39          & 97.70               & 87.50              & 93.58          \\
                                               & CenterPoint                        & 73.79                                              & 11.95                                             & 41.74          & 93.88               & 77.21              & 86.80          \\ \cmidrule(l){2-8}
                                               & \textit{average}                   & 80.48                                              & 45.41                                             & 63.78          & 94.82               & 80.52              & 88.79          \\ \midrule
            \multirow{4}{*}{DSOR~\cite{kurup2021dsor}}       & PointPillars                        & 62.50                                              & 34.96                                             & 51.27          & 87.49               & 59.81              & 75.97          \\
                                               & SECONDNet                          & 95.15                                              & 80.99                                             & 89.59          & 91.93               & 78.27              & 86.63          \\
                                               & CenterPoint                        & 51.35                                              & 10.64                                             & 30.30          & 84.26               & 65.61              & 76.59          \\ \cmidrule(l){2-8}
                                               & \textit{average}                   & 69.67                                              & 42.20                                             & 57.05          & 87.89               & 67.90              & 79.73          \\ \midrule
            \multirow{4}{*}{Particle-UNet~\cite{stanislas2021airborne}}     & PointPillars                        & 76.82                                              & 39.78                                             & 62.57          & 94.34               & 77.28              & 88.50          \\
                                               & SECONDNet                          & 99.05                                              & 78.31                                             & 91.54          & 99.18               & 86.85              & 93.81          \\
                                               & CenterPoint                        & 85.55                                              & 37.33                                             & 67.62          & 94.57               & 75.66              & 87.42          \\ \cmidrule(l){2-8}
                                               & \textit{average}                   & 87.14                                              & 51.81                                             & 73.91          & 96.03               & 79.93              & 89.91          \\ \midrule
            \multirow{4}{*}{WeatherNet~\cite{heinzler2020cnn}}        & PointPillars                        & 82.02                                              & 42.32                                             & 65.97          & 94.22               & 77.24              & 88.37          \\
                                               & SECONDNet                          & 98.80                                              & 79.96                                             & 91.62          & 98.47               & 86.83              & 93.30          \\
                                               & CenterPoint                        & 88.59                                              & 37.68                                             & 69.94          & 95.40               & 76.65              & 88.57          \\ \cmidrule(l){2-8}
                                               & \textit{average}                   & \textbf{89.80}                                     & \textbf{53.32}                                    & \textbf{75.85} & 96.03               & 80.24              & 90.08          \\ \midrule
            \multirow{4}{*}{Particle-VoxelNet~\cite{stanislas2021airborne}} & PointPillars                        & 82.84                                              & 40.93                                             & 66.45          & 94.42               & 77.18              & 87.27          \\
                                               & SECONDNet                          & 99.35                                              & 84.97                                             & 93.54          & 99.16               & 87.72              & 94.72          \\
                                               & CenterPoint                        & 79.64                                              & 12.51                                             & 44.94          & 95.65               & 78.54              & 89.11          \\ \cmidrule(l){2-8}
                                               & \textit{average}                   & 87.27                                              & 46.14                                             & 68.31          & \textbf{96.41}      & \textbf{81.15}     & \textbf{90.37} \\ \bottomrule
        \end{tabular}
    }
    \label{tab:different_filtering_nets}
\end{table*}
\textbf{Object Detection.}
We use three popular object detectors, PointPillars~\cite{lang2019pointpillars}, SECOND~\cite{yan2018second}, and CenterPoint~\cite{yin2021center}, to evaluate the performance of our framework.
All the detectors are trained on the NuScenes dataset using the implementation provided by OpenPCDet~\cite{openpcdet2020}.
As data augmentation, we use random axis flip, random scaling, and random rotation.
We use a single-scan input (no sweep accumulation) for training and testing. 
Although both NuScenes and SemanticSpray use similar $32$-layers LiDAR sensors, there is still a domain gap between the two datasets (e.g. sensor placement).
To account for this, we select a small set of LiDAR scans ($112$) from the SemanticSpray that do not contain spray and use them along with $25\%$ of the NuScenes training set to fine-tune the vanilla detectors. 
As optimization parameters for the fine-tuning, we use a learning rate of $10^{-4}$, Adam optimizer~\cite{Kingma2014AdamAM}, weight decay of $0.01$, and train for $5$ epochs.
In addition, we use copy-paste augmentation of cars and pedestrians extracted from the entire NuScenes dataset.
More information about this augmentation can be found in~\cite{openpcdet2020}.
As the domain gap effects are less pronounced in the fine-tuned models, their evaluation allows us to better isolate the effect of spray on object detection performance. 

\textbf{Evaluation Metrics.}
We use the 3D Average Precision (AP) metric, which is commonly used to evaluate 3D object detection performance. 
Similar to~\cite{mao2021one}, we provide results for detections at different ranges, namely $0$-\SI{25}{\meter} and $>\SI{25}{\meter}$. 
Additionally, we report the 3D AP for the entire detection range (overall).

\subsection{Results}
\begin{figure*}[t!]
    \centering
        \includegraphics[width=1\textwidth]{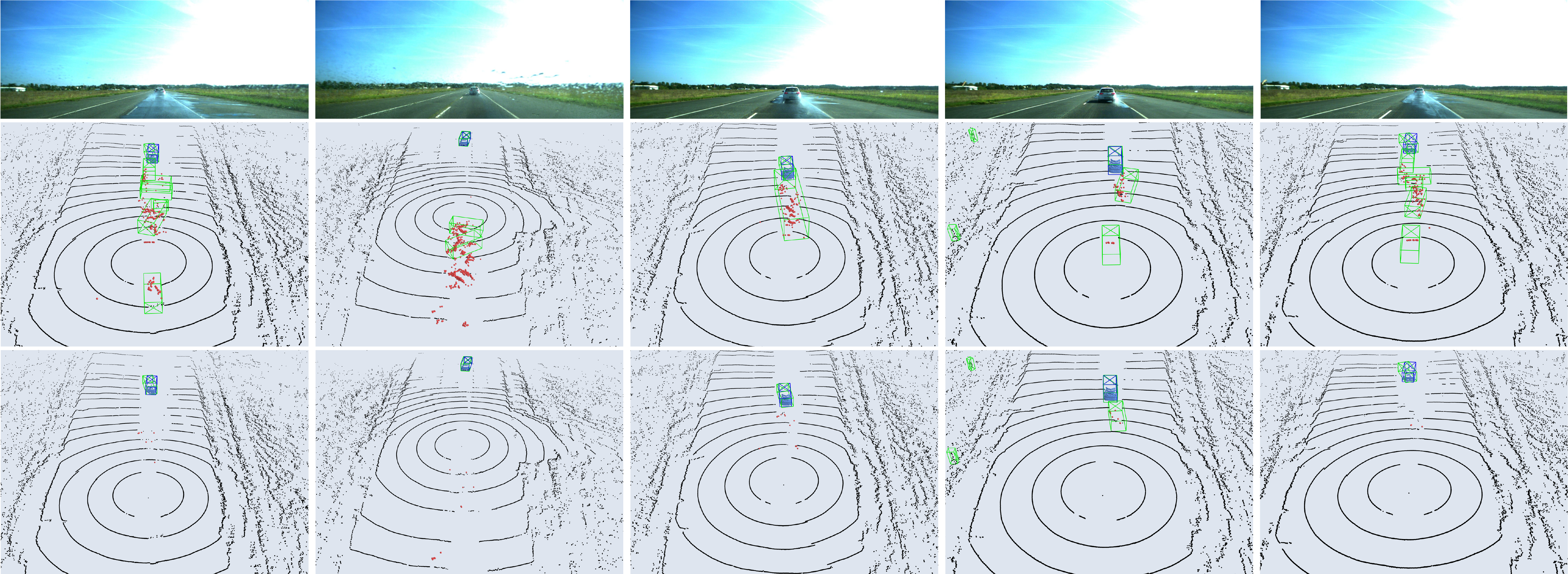}
    \caption{
Qualitative results of the proposed adverse weather filtering approach. The top images show the camera image (used for visualization only). 
The middle figures show qualitative results of SECOND without the point cloud preprocessing.
The bottom figures represent instead the results when using AWNet for adverse weather filtering.
We show spray points in red and vehicle points in light blue. We use blue boxes to represent the ground truth and green for object detections. }    \label{Fig:results}\end{figure*}

\textbf{Adverse Weather Filtering.}
In Table~\ref{tab:weather_filter_and_radar} we report the object detection results using the unprocessed point cloud as input and the filtered point cloud using AWNet.
We see that the adverse weather filtering improves the 3D AP by $5.67\%$ and $2.4\%$ points over the whole detection range for the vanilla and fine-tuned detectors respectively. 
The filtering is particularly effective in the $0$-\SI{25}{\meter} range, where most of the spray points from both the ego and leading vehicle are present.
In Fig.~\ref{Fig:results} we show some qualitative results of the adverse weather filtering on 3D object detection.
When comparing the vanilla and fine-tuned detectors without preprocessing, we see that there is a large gap in performance, highlighting the problems of domain adaptation outside of the spray effect.

\textbf{Radar Target Filtering.}
In Table~\ref{tab:weather_filter_and_radar} we also report the results for the radar-based detection filtering.
When only the radar targets are used to filter the object predictions, we see a large gain in performance compared to the baseline results.
For example, in the overall range, we see an improvement in 3D AP of $17.84\%$ and $3.49\%$ points for the vanilla and fine-tuned detectors respectively.
As radar detections are less affected by adverse weather conditions and the SemanticSpray dataset contains only a single target vehicle, the majority of false positive predictions are filtered out, resulting in high performance for the vanilla detectors as well.
Finally, when using both the adverse weather and radar target filters, we see an additional improvement in performance for the vanilla and fine-tuned models.

\textbf{Different Adverse Weather Filters.}
Table~\ref{tab:different_filtering_nets} shows the results of the adverse weather filtering using different methods. 
All the learning-based methods improve the 3D AP performance of both the vanilla and fine-tuned detectors.
Only when the DSOR filter is applied, a large decrease in performance is observed.
DSOR is designed to filter out snowfall points, which are usually detected as dense clusters around the sensor.
Instead, spray points are detected as scattered clusters at greater distances from the sensor, making the use of local neighbors less effective.  
Compared to AWNet, the CNN-based methods (Particle-UNet and WeatherNet) achieve better performance for the vanilla detectors.
Instead, for the fine-tuned detectors, AWNet has the higher performance of all the methods tested. 
As reported in~\cite{Piroli2023EnergybasedDO}, the inference time for the tested methods are \SI{71.50}{\ms} DSOR, \SI{4.32}{\ms} Particle-UNet, \SI{4.18}{\ms} WeatherNet, \SI{409.14}{\ms} Particle-VoxelNet and \SI{15.37}{\ms} AWNet.
When using fast object detectors like PointPillars (\SI{16.13}{\ms} inference time~\cite{lang2019pointpillars}), the use of filtering methods such as AWNet,  Particle-UNet, and WeatherNet still allows for real-time performance considering the usual $10$-$20$\SI{}{\hertz} sampling frequency of LiDAR sensors.

\subsection{Ablation Studies}
\begin{table}[t!]
    \centering
    \caption{Impact of the threshold $\tau$ used to classify points as \textit{valid} or \textit{noise}.
        The results refer to the 3D AP metric measured at different distances and on the overall detection range. 
        Values are in percentage. Bold numbers represent the best results.}
    \resizebox{0.9\columnwidth}{!}{%
        \begin{tabular}{@{}clccc@{}}
            \toprule
            \multirow{2}{*}{$\tau$} & \multirow{2}{*}{\textbf{Detector}} & \multicolumn{3}{c}{\textbf{3D AP w/ fine-tuning}}                                       \\ \cmidrule(l){3-5}
                                    &                                    & \textbf{$0$-\SI{25}{\meter}}                      & $>\SI{25}{\meter}$ & overall        \\ \midrule
            \multirow{4}{*}{$90\%$ TPR} & PointPillars                        & 93.86                                             & 77.03              & 86.89          \\
                                    & SECOND                             & 99.10                                             & 85.85              & 94.72          \\
                                    & CenterPoint                        & 94.75                                             & 77.45              & 87.99          \\ \cmidrule(l){2-5}
                                    & \textit{average}                   & 95.91                                             & 80.11              & 89.87          \\ \midrule
            \multirow{4}{*}{$95\%$ TPR} & PointPillars                        & 94.27                                             & 77.07              & 87.19          \\
                                    & SECOND                             & 99.31                                             & 85.85              & 94.83          \\
                                    & CenterPoint                        & 95.31                                             & 77.91              & 88.60          \\ \cmidrule(l){2-5}
                                    & \textit{average}                   & 96.30                                             & 80.28              & 90.21          \\ \midrule
            \multirow{4}{*}{$99\%$ TPR} & PointPillars                        & 95.02                                             & 77.20              & 88.64          \\
                                    & SECOND                             & 99.43                                             & 85.64              & 94.83          \\
                                    & CenterPoint                        & 96.15                                             & 79.38              & 90.11          \\ \cmidrule(l){2-5}
                                    & \textit{average}                   & \textbf{96.87}                                    & \textbf{80.74 }    & \textbf{91.19} \\ \midrule
        \end{tabular}
    }
    \label{tab:AWN_tpr_treshold_ablation}
\end{table}
\textbf{Noise Classification Threshold.}
In Table~\ref{tab:AWN_tpr_treshold_ablation} we report the effect of different decision thresholds $\tau$~\eqref{eq:decision_rule} on performance. 
For outlier-based methods such as AWNet, it is common to set a decision threshold based on a desired TPR percentage~\cite{liu2020energy,du2022vos,Piroli2023EnergybasedDO}. 
We see that this choice can directly affect the performance of 3D object detection.
For example, although $\tau$ at $95\%$ TPR will filter out more spray points than $\tau$ at $99\%$ TPR, the performance of the former is lower because some vehicle points are also filtered out.
In addition, important context clues such as road points may be lost when using higher filtering thresholds. 

\begin{table}[t!]
    \centering
    \caption{Impact of the box padding parameter $\gamma$ when using radar targets for false positive detection filtering. 
        The results refer to the 3D AP metric measured at different distances and on the overall detection range. 
        Values are in percentage. Bold numbers represent the best results.}
    \resizebox{0.95\columnwidth}{!}{%
        \begin{tabular}{@{}clccc@{}}
            \toprule
            \multirow{2}{*}{\textbf{Box Padding}} & \multirow{2}{*}{\textbf{Detector}} & \multicolumn{3}{c}{\textbf{3D AP w/ fine-tuning}}                                       \\ \cmidrule(l){3-5}
                                                  &                                    & \textbf{$0$-\SI{25}{\meter}}                   & $>\SI{25}{\meter}$ & overall        \\ \midrule
            \multirow{4}{*}{\SI{0}{\meter}}       & PointPillars                        & 89.58                                          & 72.76              & 82.89          \\
                                                  & SECOND                             & 95.67                                          & 82.33              & 91.23          \\
                                                  & CenterPoint                        & 93.59                                          & 78.65              & 88.83          \\ \cmidrule(l){2-5}
                                                  & \textit{average}                   & 92.95                                          & 77.91              & 87.65          \\ \midrule
            \multirow{4}{*}{\SI{0.5}{\meter}}     & PointPillars                        & 95.51                                          & 78.56              & 88.77          \\
                                                  & SECOND                             & 97.68                                          & 86.26              & 95.16          \\
                                                  & CenterPoint                        & 97.58                                          & 82.54              & 92.79          \\ \cmidrule(l){2-5}
                                                  & \textit{average}                   & 96.92                                          & 82.46              & 92.24          \\ \midrule
            \multirow{4}{*}{\SI{1}{\meter}}       & PointPillars                        & 95.54                                          & 78.62              & 88.81          \\
                                                  & SECOND                             & 99.66                                          & 86.29              & 95.19          \\
                                                  & CenterPoint                        & 97.61                                          & 82.62              & 92.84          \\ \cmidrule(l){2-5}
                                                  & \textit{average}                   & \textbf{97.60}                                 & 82.51              & \textbf{92.28} \\ \midrule
            \multirow{4}{*}{\SI{1.5}{\meter}}     & PointPillars                        & 95.54                                          & 78.66              & 88.82          \\
                                                  & SECOND                             & 99.66                                          & 86.30              & 95.19          \\
                                                  & CenterPoint                        & 97.61                                          & 82.63              & 92.84          \\ \cmidrule(l){2-5}
                                                  & \textit{average}                   & \textbf{97.60}                                 & \textbf{82.53}     & \textbf{92.28} \\ \bottomrule
        \end{tabular}
    }
    \label{tab:radar_box_padding}
\end{table}
\textbf{Box Padding.}
In Table~\ref{tab:radar_box_padding} we show the effect of different padding values on 3D AP.
When applying $\SI{0}{\meter}$ padding we observe lower performance compared to the non-processed method.
This is because radar targets for the leading vehicle are usually located on the back, which makes the filtering approach sensible to size and orientation estimation.
By adding $\SI{0.5}{\meter}$ padding, we see higher performance than non-processed input.
A small improvement is instead seen when increasing the padding to $\SI{1.0}{\meter}$ and $\SI{1.5}{\meter}$.
However, in dense and cluttered scenarios where multiple detections are clustered together, larger padding values may lead to overlapping of detections.

\subsection{Discussion}
The experiments presented in the previous sections show promising results in improving the robustness of 3D object detectors in rainy conditions.
However, the small amount of available labeled data containing road spray limits the possible testing of this effect.
The Road Spray and SemanticSpray datasets are a first step towards a better understanding of the effect of spray on the perception of autonomous driving systems. 
Nevertheless, the scenes in the datasets are limited to a single lead vehicle in an uncluttered environment.
In real-world applications, more complex scenes may occur, making the presented filtering approaches less effective.
Finally, the use of radar sensors can be further improved by using more sophisticated rules for filtering or directly integrating the radar targets in the LiDAR object detector training.
\section{Conclusion}
In this paper, we have presented a simple but effective framework for improving the robustness of 3D object detectors trained on good weather data against road spray.
The framework can be used in both a LiDAR-only setup or it can be extended to include measurements from radar sensors.
Our idea consists of using a state-of-the-art network~\cite{Piroli2023EnergybasedDO} to reliably filter out adverse weather points  in a LiDAR point cloud.
Using the filtered data, we show that the object detection performance of PointPillars, SECOND, and CenterPoint is improved.
Additionally, we use the inherent robustness of radar sensors to adverse weather to further filter out false positive object detections.
We test our framework on the SemanticSpray dataset~\cite{Piroli2023EnergybasedDO}, which consists of real-world highway-like scenarios containing road spray. 
In future work, we will explore the use of our framework in more complex scenarios and the inclusion of additional sensor modalities.

\bibliographystyle{IEEEtran}
\bibliography{mybib}

\end{document}